\pdfoutput=1

\documentclass[10pt,twocolumn,letterpaper]{article}

\usepackage{capt-of}
\usepackage{graphicx}
\usepackage{wrapfig}
\usepackage{indentfirst}
\usepackage{amsmath}
\usepackage{multirow} 
\usepackage{array} 
\usepackage{tabularx}
\newcolumntype{Y}{>{\centering\arraybackslash}X}
\usepackage[pagenumbers]{cvpr} 

\usepackage{colortbl}
\newcommand{\best}[1]{\cellcolor[HTML]{FD6864}{#1}}
\newcommand{\second}[1]{\cellcolor[HTML]{FFFC9E}{#1}}

\usepackage{placeins}

\definecolor{cvprblue}{rgb}{0.21,0.49,0.74}
\usepackage[pagebackref,breaklinks,colorlinks,allcolors=cvprblue]{hyperref}

\def\paperID{10782} 
\def\confName{CVPR}
\def\confYear{2026}

\title{ProgressiveAvatars: Progressive Animatable 3D Gaussian Avatars}

\author{Kaiwen Song \quad Jinkai Cui \quad Juyong Zhang\textsuperscript{*}\\
University of Science and Technology of China}

\begin{document}

\twocolumn[{%
\renewcommand\twocolumn[1][]{#1}%
\renewcommand{\thefootnote}{\fnsymbol{footnote}}
\maketitle 
\renewcommand{\thefootnote}{\arabic{footnote}}
\begin{center}
    \captionsetup{type=figure}
    \includegraphics[width=\textwidth]{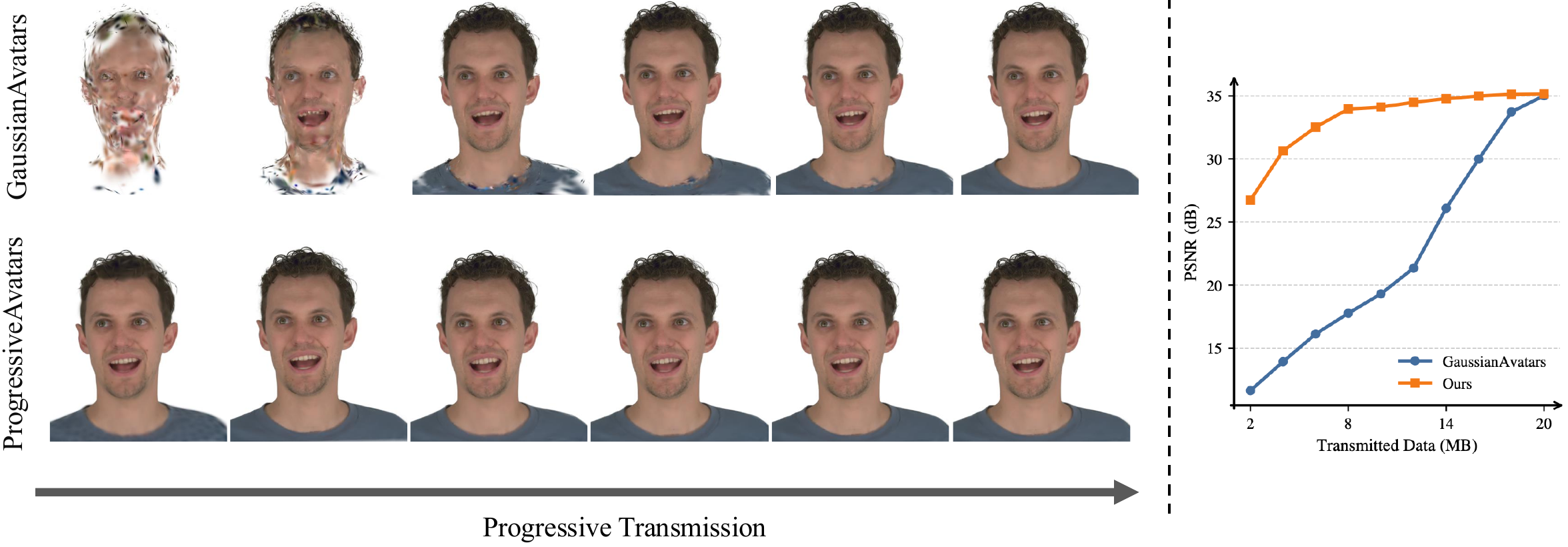} 
    \caption{\textbf{ProgressiveAvatars} is a novel progressive representation that supports
  adaptive rendering quality of 3D Gaussian avatars under bandwidth or compute
  constraints. Qualitative (left) and quantitative (right) results demonstrate that ProgressiveAvatars rapidly attains high quality and continues to refine the avatar as more data arrives, whereas GaussianAvatars~\cite{hu2024gaussianavatar} only becomes usable once nearly
    the entire asset has been transmitted.}
    \label{fig:teaser}
\end{center}
}] 

\begingroup
\renewcommand{\thefootnote}{\fnsymbol{footnote}}
\footnotetext[1]{Corresponding author.}
\endgroup

\begin{abstract}
In practical real-time XR and telepresence applications, network and computing resources fluctuate frequently. Therefore, a progressive 3D representation is needed. To this end, we propose ProgressiveAvatars, a progressive avatar representation built on a hierarchy of 3D Gaussians grown by adaptive implicit subdivision on a template mesh. 3D Gaussians are defined in face‑local coordinates to remain animatable under varying expressions and head motion across multiple detail levels. The hierarchy expands when screen-space signals indicate a lack of detail, allocating resources to important areas. Leveraging importance ranking, Progressive-Avatars supports incremental loading and rendering, adding new Gaussians as they arrive while preserving previous content, thus achieving smooth quality improvements across varying bandwidths. ProgressiveAvatars enables progressive delivery and progressive rendering under fluctuating network bandwidth and varying compute and memory resources. Project page: \href{https://ustc3dv.github.io/ProgressiveAvatars/}{ustc3dv/ProgressiveAvatars}
\vspace{-2em}
\end{abstract}
    
\vspace{-2em}
\section{Introduction}
\label{sec:intro}

High-fidelity, real-time rendered head avatars are crucial for immersive interaction, visual communication, and digital human creation. In dynamic multi-user scenarios like Social VR, transmitting high-fidelity avatars as conventional static assets causes severe start-up latency and bandwidth spikes, breaking immersion by forcing users to wait for complete downloads before any rendering can begin. This calls for a progressive 3D representation that provides an animatable avatar capable of minimal start-up latency and continuously refining details as more data arrives.  Due to its high fidelity and efficient rendering, 3D Gaussian Splatting~\cite{kerbl3Dgaussians,wu2024recent} and its variants are gradually becoming the mainstream explicit representation. 
Recent advancements have rapidly expanded across high-fidelity head and body avatars~\cite{xiang2024flashavatar,wang2025mega,qian2024gaussianavatars,hu2024gaussianavatar,zielonka2023drivable,peng2025pica}, real-time SLAM~\cite{peng2024rtgslam,peng2025gpsslam}, and advanced rendering techniques~\cite{ye2024gsdr,lee2025gaussiansurfel}. However, existing 3DGS-based avatar methods rarely support progressive, streamable 3D representations. They lack incremental loading mechanisms, failing to smoothly accumulate details with increasing input bandwidth while maintaining controllability and temporal stability. Therefore, current 3D Gaussian-based digital human representations are unsuitable for these latency-sensitive scenarios.

Constructing a progressive 3D avatar faces several challenges. First, the constructed model must remain animatable across varying transmission budgets. Second, it must maintain temporal stability across various dynamic scenarios under different expressions and head movements. Directly increasing the number of Gaussians can generally improve modeling accuracy, but it heavily increases transfer size, loading times, and instantiation overhead, while excessively restricting capacity compromises modeling accuracy. Prior works~\cite{yan2025architecthead,dongye2024lodavatar} typically employ a uniform capacity expansion approach, such as increasing UV resolution or uniformly subdividing the template mesh. This may over-refine smooth areas while under-refining high-frequency areas, resulting in wasted resources. Moreover, existing methods~\cite{liu2024citygaussian,dongye2024lodavatar,fan2024lightgaussian,ren2024octree} that attempt to offer multiple detail levels typically rely on a discrete LOD paradigm. This requires generating and storing multiple independent copies of the same avatar for different quality budgets, leading to severe storage redundancy and rigid, latency-inducing asset switching. As a result, details cannot be added incrementally, and true progressive transmission and rendering is not supported.

Based on these practical needs and the shortcomings of existing representations, we propose a progressive head avatar representation utilizing adaptive implicit subdivision and continuous detail accumulation. Unlike discrete LOD pipelines that rigidly switch between redundant models, our single, unified asset allows any received subset of Gaussians to be rendered immediately at arbitrary transmission percentages. Built upon the FLAME model~\cite{FLAME:SiggraphAsia2017} to provide a base animatable structure, we construct a mesh-anchored subdivision hierarchy. By binding 3D Gaussians to the local coordinates of each triangle, the avatar remains animatable under mesh deformation and structurally consistent as details incrementally accumulate. Furthermore, rather than using uniform subdivision, we rely on screen-space gradient signals to adaptively refine only detail-rich regions, optimizing resource allocation. Ultimately, as more data streams in, newly arrived Gaussians are seamlessly merged without discarding prior content, enabling true progressive rendering and eliminating the need for cumbersome full-model replacements. In summary, our main contributions are:

\begin{itemize}
    \item We propose ProgressiveAvatars, a novel progressive 3D avatar representation. It shifts the paradigm from discrete, redundant LOD switching to a single, continuous streamable asset. It grows progressively through adaptive implicit subdivision on a mesh-anchored 3D Gaussian hierarchy, remaining animatable across all streaming stages and recovering fine-scale details via adaptive refinement.
    \item We achieve incremental loading and progressive rendering through gradual activation. Each streaming increment adds new 3D Gaussians to the previous content, allowing a coarse but animatable avatar to emerge instantly, with visual quality smoothly improving as bandwidth allows.
    \item Extensive experiments demonstrate that ProgressiveAvatars can rapidly generate high-quality rendering results as data streams in, and its finest modeling results achieve modeling quality comparable to state-of-the-art methods.
\end{itemize}

\section{Related Work}
\label{sec:related}

\begin{figure*}[t]
    \centering

        \includegraphics[width=\textwidth]{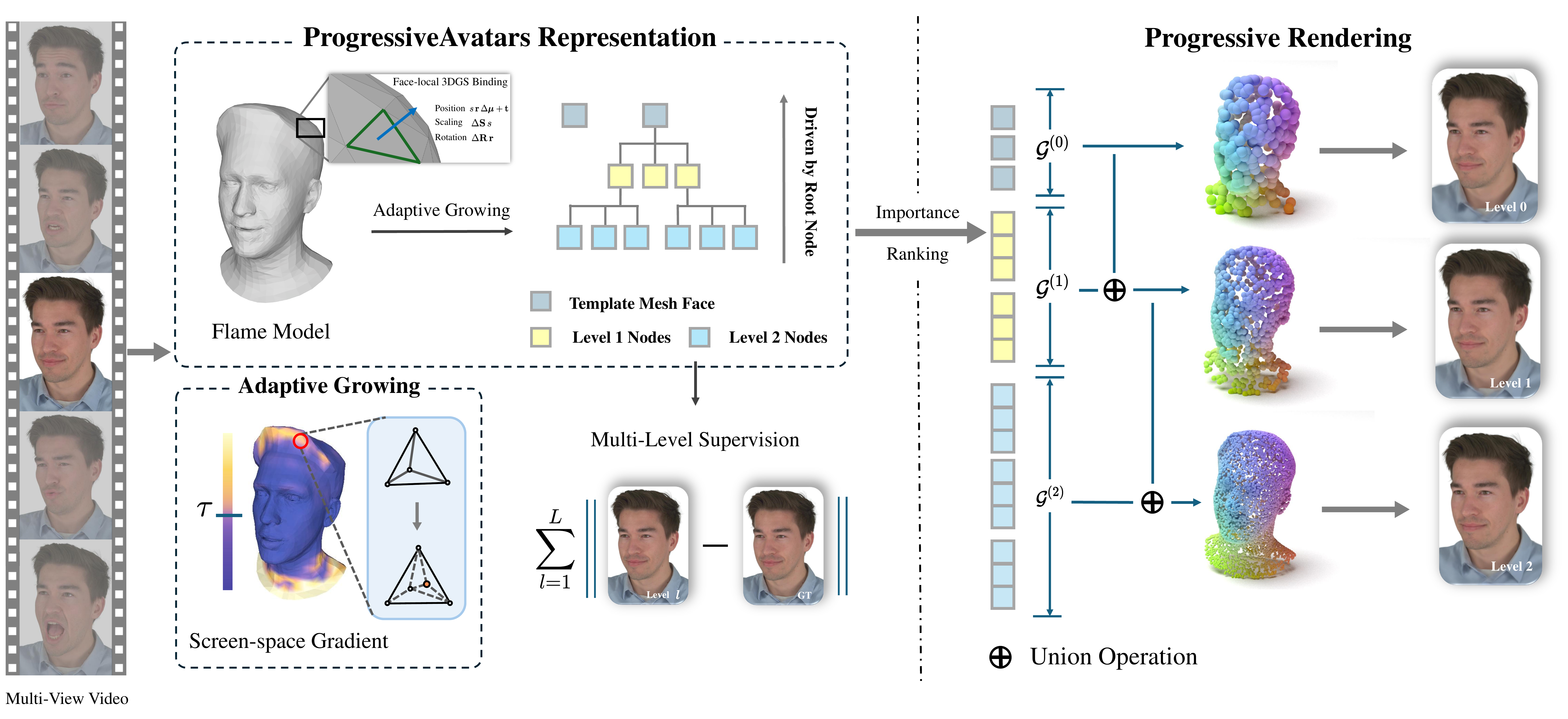}%

  \caption{\textbf{Overview.} We take head video as input then recover a tracked FLAME mesh sequence. We bind 3D Gaussians to the
  local coordinate frame of each FLAME face. During training, screen-space gradients of the Gaussians drive implicit subdivision of the template mesh across multiple levels, yielding a triangle face forest. At rendering time, we precompute per-face importance score and progressively transmit and render the corresponding Gaussians in decreasing order of importance.}
    \label{fig:overview}
\end{figure*}

\subsection{Animatable Head Avatars}
Classical head avatar pipelines fit a morphable mesh with fixed topology to drive animation. Such templates are robust for tracking and retargeting, but they inherently miss delicate geometry and struggle to reproduce complex, view‑dependent appearance~\cite{fu2008realtime,hogue2007automated,ichim2015dynamic,leung2000realistic,lombardi2018deep}. To increase realism, many works augment meshes with implicit fields like NeRF~\cite{mildenhall2021nerf}.  NerFace~\cite{gafni2021nerface} learns deformation fields to animate heads, but can produce floaters and struggles to match precise target expressions. HeadNeRF~\cite{hong2021headnerf} builds a parametric head NeRF and uses lightweight 2D neural rendering for efficiency. INSTA~\cite{zielonka2022insta} maps query points to a canonical space via nearest‑triangle search on a FLAME~\cite{FLAME:SiggraphAsia2017} mesh and couples this with InstantNGP~\cite{mueller2022instant} for fast rendering. NeRFBlendShape~\cite{gao2022reconstructing} models dynamics by blending hash‑grid features conditioned on 3DMM parameters. PointAvatar~\cite{Zheng_2023_CVPR} represents heads as colored, deformable points and is effective for animation.

Recently, 3D Gaussian Splatting  has emerged as a compelling explicit representation for head avatars, combining high fidelity with real‑time rendering~\cite{kerbl3Dgaussians}. GaussianAvatars~\cite{qian2024gaussianavatars} rigs Gaussians to tracked FLAME meshes to obtain controllable, high quality head animation. Splatting Avatar~\cite{shao2024splattingavatar}, embed trainable Gaussians on a template mesh, initializing them by random
surface sampling and refining them through a walking on triangles strategy. FlashAvatar~\cite{xiang2024flashavatar} presents a lightweight mesh‑embedded Gaussian field with residual offsets in UV‑space. MeGA~\cite{wang2025mega} mixes meshes and Gaussians by regions to leverage the strengths of each representation, achieving superior quality and enabling head editing after a short optimization stage. However, under tight bandwidth or compute budgets, these methods lack mechanisms to adapt level of detail at inference time, making them difficult to stream or to scale gracefully.

\subsection{Level of Detail}
Level of Detail  traditionally balances scene complexity against performance in interactive graphics. A large body of work integrates LoD into classic pipelines~\cite{lodclark1976hierarchical,lodcrassin2009gigavoxels,lodduchaineau1997roaming,lodguthe2002interactive,lodlindstrom2001visualization,lodluebke2003level,lodprogressive,losasso2004geometry} to reduce memory traffic, streamline rendering, and preserve responsiveness. In neural implicit reconstruction, NGLoD~\cite{takikawa2021lod} organizes features in a sparse voxel octree whose depth indexes detail, and it enables continuous LOD and
  real‑time SDF rendering by interpolating across levels and traversing the octree sparsely. Takikawa et al.\cite{takikawa2022variable}
   compress multi‑resolution
  feature grids with a vector‑quantized dictionary, which reduces memory by up to orders of magnitude and supports hierarchical streaming at variable
  bitrate. BungeeNeRF~\cite{xiangli2022bungeenerf} trains progressively for extreme multi‑scale scenes. It first fits distant views with a shallow base block, then appends new
  blocks as training proceeds while activating high‑frequency channels in the positional encoding to reveal finer details. Tri‑MipRF~\cite{hu2023tri}  prefilters
  the 3D feature space into three orthogonal mipmaps and performs cone‑casting for area sampling, which delivers anti‑aliased renderings with fast
  reconstruction. LoD‑NeuS~\cite{zhuang2023anti} adopts a multi‑scale tri‑plane representation and aggregates features within a conical frustum along each ray, recovering 
  high‑frequency surface detail while suppressing aliasing.

With advancements in 3DGS, researchers have started exploring modeling different LOD in explicit 3D Gaussian scenes~\cite{fan2024lightgaussian,liu2024citygaussian,fischer2024dynamic,ren2024octree,kerbl2024hierarchical,kulhanek2025lodge,song2024city,song2025structuredfield}. CityGaussian compresses trained large‑scale 3DGS at multiple ratios by  LightGaussian~\cite{fan2024lightgaussian} to obtain LoD variants. 4DGF~\cite{fischer2024dynamic} integrated 3D Gaussians as an efficient geometry scaffold while utilizing neural fields as a compact and flexible appearance model. Octree‑GS~\cite{ren2024octree} organizes multi‑scale Gaussians in an octree and selects levels dynamically by viewing distance and projected texture frequency, achieving  rendering at stable frame rates. Hierarchical 3D Gaussians~\cite{kerbl2024hierarchical} and LODGE~\cite{kulhanek2025lodge} choose levels based on the 2D projected scale of Gaussians for street‑scale scenes, enabling LoD training and rendering. These methods typically pre‑materialize all levels and, at runtime, select or switch levels from camera distance or primitive projected are. In contrast to prior work that mainly targets rendering throughput, we focus on streaming a 3DGS avatar. We aim get a usable head avatar appears quickly at a coarse LoD, and the remaining Gaussians are incrementally transmitted to refine quality with controllable compute and bandwidth.

A closely related line of work includes LoDAvatar~\cite{dongye2024lodavatar} and ArchitectHead~\cite{yan2025architecthead}. Neither method supports progressive rendering, since both require switching rendering resources. LoDAvatar embeds 3D Gaussian splats into a template mesh and performs uniform subdivision across levels, which tends to over‑refine uninformative regions and waste computation. To partially address this issue, it applies a hand‑crafted mask at render time to selectively densify specified areas and regulate the number of Gaussians. In contrast, our approach grows a multi‑level representation through adaptive optimization, achieving a more favorable trade‑off between visual fidelity and primitive count. ArchitectHead parameterizes Gaussians in a 2D UV feature space and defines a UV feature field composed of multi‑level learnable maps. A neural decoder then maps the latent features to 3D Gaussian attributes for rendering. Although this design offers a tunable balance between fidelity and efficiency, its network‑based formulation causes the frame rate to deteriorate rapidly as the number of Gaussians increases, which in turn limits  applicability in production settings when compared with purely explicit pipelines.

\section{Method}
As shown in~\cref{fig:overview}, ProgressiveAvatars takes video records of a human head as input. For each time step $t$, we fit a FLAME~\cite{FLAME:SiggraphAsia2017} mesh $M_t=(\mathbf{V}_t,\mathbf{F})$ using a photometric multi-view tracker, where the vertex positions $\mathbf{V}_t$ vary with expressions and poses while the mesh topology $\mathbf{F}$ remains fixed over time. Leveraging this topological consistency, we build a progressive representation by implicitly subdividing the template topology to obtain a hierarchy $\{\mathbf{F}^{(\ell)}\}_{\ell=0}^{L}$, and we associate 3D Gaussians with faces at each level $\ell$, where $\mathcal{G}^{(\ell)}=\bigcup_{f\in\mathbf{F}^{(\ell)}}\mathcal{G}^{(\ell)}_f$. We render images from $\mathcal{G}^{(\ell)}$ using a differentiable Gaussian rasterizer and supervise them with multi-view ground-truth images. This yields an animatable head avatar that supports progressive refinement. During training, the hierarchy grows on the fly via implicit subdivision guided by screen-space signals so detail concentrates where needed. During inference, we realize progressive rendering through incremental loading. Newly available Gaussians from finer levels are added while previously loaded content remains unchanged, and quality improves smoothly as more data arrives.

\subsection{ProgressiveAvatars Representation}
\label{sec:progressive}
Our goal is to construct a progressive representation for head avatars. We grow a hierarchical tree on each template mesh face through implicit subdivision and bind 3D Gaussians to faces at each level. Gaussians are defined in the corresponding face-local frame so they remain stably animatable under varying expressions and poses and preserve identity across levels. At the coarsest level, base bindings cover all template faces, providing a complete global avatar before any fine-level refinement arrives.

\noindent\textbf{Implicit Subdivision.}\quad  Starting from $\mathbf{F}^{(0)}=\mathbf{F}$, we apply recursive triangle subdivisions to obtain $\{\mathbf{F}^{(\ell)}\}_{\ell=0}^{L}$. For a parent face $f=(i,j,k)\in\mathbf{F}^{(\ell)}$, a new child vertex $\mathbf{p}$ is created by barycentric interpolation of the parent vertices $\{\mathbf{v}_i,\mathbf{v}_j,\mathbf{v}_k\}$:
\begin{equation}
\mathbf{p} = \beta_1\,\mathbf{v}_i + \beta_2\,\mathbf{v}_j + \beta_3\,\mathbf{v}_k.
\end{equation}
The barycentric coefficients $\boldsymbol{\beta}=(\beta_1,\beta_2,\beta_3)$ are initialized to $(\tfrac{1}{3},\tfrac{1}{3},\tfrac{1}{3})$ and are softly optimized during training under the simplex constraint $\beta_i\ge 0$ and $\sum_i \beta_i=1$. This lets $\mathbf{p}$ move within the triangle and induces different effective scales and shapes per face so the hierarchy adapts to facial regions of different sizes and structures while keeping a consistent per-level construction. Under varying expressions and poses, the positions of subdivided points are computed at each level by the same barycentric mapping. These per-face hierarchical trees collectively form a forest over the template mesh.

\noindent\textbf{Face-Local Gaussian Binding.}\quad On this hierarchy, we bind each 3D Gaussian to a triangle using a residual, face-local parameterization similar to GaussianAvatars~\cite{hu2024gaussianavatar}:

\begin{equation}
\mathbf{R} = \Delta\mathbf{R}\,\mathbf{r},\qquad
\mathbf{S} = \Delta\mathbf{S}\,s,\qquad
\boldsymbol{\mu} = s\,\mathbf{r}\,\Delta\boldsymbol{\mu} + \mathbf{t},
\end{equation}
where $\Delta\mathbf{R}$, $\Delta\boldsymbol{\mu}$, and $\Delta\mathbf{S}$ are trainable residuals, $\mathbf{r}$ is a rotation aligned with the triangle normal, $\mathbf{t}$ is the triangle centroid, and $s$ is a face scale computed as the mean of the three edge lengths. This binding makes Gaussians co-move with their binding faces under varying expressions and poses and preserves consistent appearance across levels.

\noindent\textbf{Progressive Rendering.}\quad ProgressiveAvatars is a natural representation for progressive transmission and rendering. After training we sort the face hierarchy and assign each face at every level an importance score $W_i$. The score determines the order of transmission and activation and is obtained from aggregated screen-space gradients and view statistics collected during training, as described in~\cref{sec:construction}. At rendering time, the coarsest avatar is transmitted first, followed by a stream of refinement records. Each record carries the barycentric coordinates $\boldsymbol{\beta}$ of the subdivision point and the residual parameters $\Delta\mathbf{R}$, $\Delta\mathbf{S}$, and $\Delta\boldsymbol{\mu}$ for the newly introduced 3D Gaussians. The receiver incrementally adds these Gaussians as records arrive and drives them with the template mesh by recursively evaluating the barycentric mapping from the root face to the target face. Crucially, this realizes continuous detail accumulation rather than discrete asset switching. This progressive rendering has the following advantages.
\begin{itemize}
  \item \textbf{Progressive transmission.} Uses a priority-ordered stream guided by learned per-face scores, enabling bandwidth-adaptive refinement and faster time to a usable avatar without retransmitting earlier content.
  \item \textbf{Continuous detail accumulation.} Delivers pop-free and temporally stable transitions via additive refinement, and supports continuous view- and content-aware adjustment through selective subdivision.
  \item \textbf{Adaptive refinement.} Thanks to our hierarchical tree with inherited parent–child order on each template face, we can realize adaptive refinement. For example, we can apply adaptive subdivision only to facial regions such as the eyes or mouth. This further focuses resources on user‑attended regions.
\end{itemize}

\subsection{ProgressiveAvatars Construction}
\label{sec:construction}
\noindent\textbf{Adaptive Subdivision.} We couple 3DGS adaptive density control with the hierarchical trees to build the structure on the fly during training. We accumulate the screen-space gradient $g_i$ only at the current finest level $\ell_{\max}$, since the finest structure most faithfully reflects whether the current representation has sufficient detail, and then grow the structure at a fixed step size $k$. Every $k$ iterations, we select leaf faces satisfying $g_i>\varepsilon$ and subdivide them. For each new child face $f'$, we attach a unique Gaussian as described in~\cref{sec:progressive}. This training–growth loop continues until the maximum level $L$ is reached for each face. This strategy grows the per-face hierarchical trees in the most demanding fine-scale regions, yielding better marginal quality at a fixed budget.

\begin{figure}[t]
    \centering
        \includegraphics[width=\columnwidth]{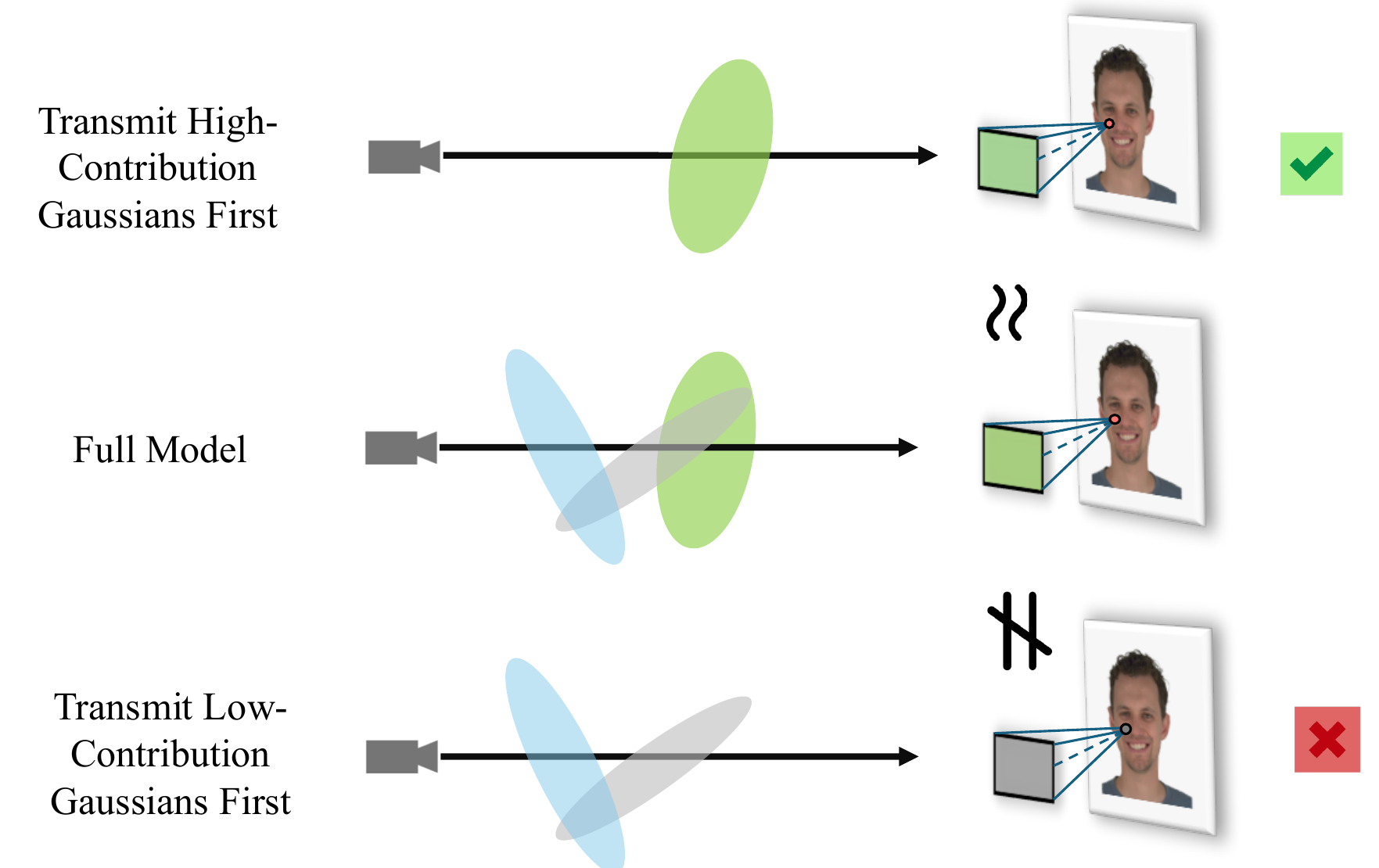}%
  \caption{%
  The center row shows the full model containing all 3D Gaussians within one level.
  Transmitting in descending importance makes early partial renderings closely match the full‑model pixel color  because dominant contributors arrive first. In contrast, sending low‑importance Gaussians first re‑normalizes partial weights and amplifies weak contributors, causing noticeable color drift from the full model. This motivates an importance‑first schedule within each level for faithful progressive rendering.}
    \label{fig:imp}
\end{figure}

\begin{figure*}[!t]
    \centering
        \includegraphics[width=\textwidth]{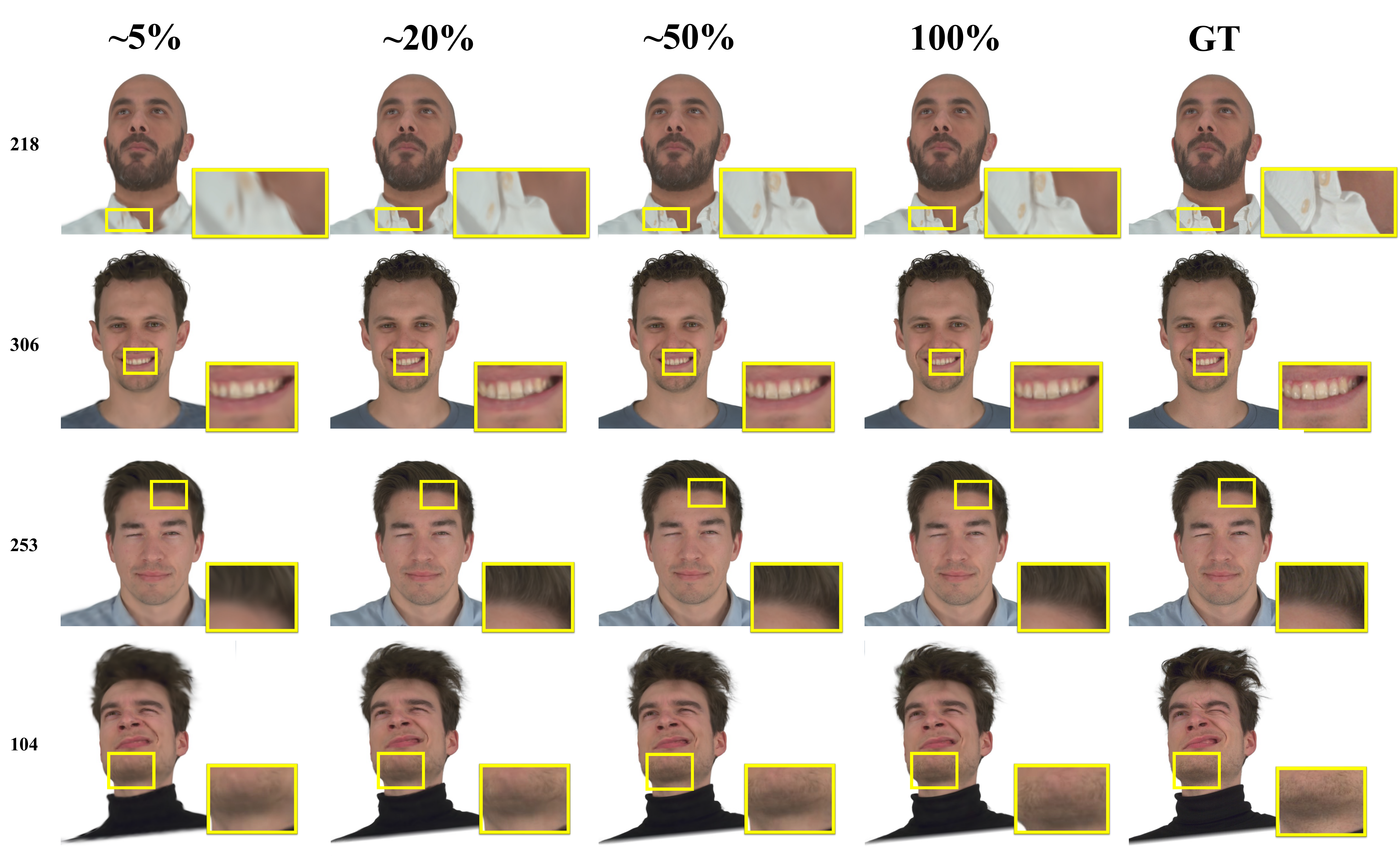}%
    \caption{\textbf{Qualitative results on NeRSemble dataset across different transmission percentages.}}
    \label{fig:lod_result}
\end{figure*}

\noindent\textbf{Importance Ranking.}  After training, we linearize the multi‑level hierarchy into a one‑dimensional 3D asset stream to enable progressive transmission. Inspired by~\cite{girish2024eagles,fan2024lightgaussian}, we compute an importance score for each face at every level and sort faces within the level accordingly.  For face $i$, we define the score as the aggregated rendering contribution of its bound Gaussians over all pixels $p$:
\begin{equation}
  W_i = \sum_{j \in \mathcal{G}_i} \sum_{p} \alpha_{j,p}\, T_{j,p},
\label{eq:score}
\end{equation}
where $\mathcal{G}_i$ is the set of Gaussians bound to face $i$, $\alpha_{j,p}$ is the per‑pixel opacity, and $T_{j,p}$ is the accumulated transmittance up to Gaussian $j$ along the viewing ray through pixel $p$. In progressive transmission or continuous streaming settings, we prioritize higher scored faces and their Gaussians into the renderer.~\cref{fig:imp} shows that  importance guided scheduling reduces color drift and suppresses artifacts, improving perceptual quality under tight bandwidth and compute.

\subsection{Training}

\noindent\textbf{Multi-Level Supervision.} We jointly supervise multiple levels to encourage cross-level consistency. Let $\mathcal{S}$ be the set of supervised levels, the photometric loss is
\begin{equation}
\mathcal{L}_{\mathrm{rgb}}=\sum_{\ell\in\mathcal{S}} w_\ell\,\Big[(1-\lambda_{s})\,\mathcal{L}_1+\lambda_s\,\mathcal{L}_{\text{ssim}}\Big],
\end{equation}
with per-level supervision weights $w_\ell$. We employ a coarse‑to‑fine optimization strategy that incrementally relaxes the upper bound on the per‑face hierarchical subdivision depth. At initialization, the depth cap is set to 1. Every 50k iterations we increase this cap and invoke adaptive subdivision to expand the forest under the updated budget, continuing until the maximum depth $D$ is reached.

\noindent\textbf{Regularization.} We further adopt the scaling loss $\mathcal{L}_{\text{scale}}$ and the position loss $\mathcal{L}_{\text{pos}}$ from~\cite{hu2024gaussianavatar} to constrain the relative position and scale with respect to its bound face, thereby preventing artifacts that occur when the 3D Gaussians move with the FLAME mesh.

Our final loss function is
\begin{equation}
    \mathcal{L}=\mathcal{L}_{\mathrm{rgb}}+\lambda_{\text {scale }} \mathcal{L}_{\text {scale }}+\lambda_{\mathrm{pos}}\mathcal{L}_{\mathrm{pos}}.
\end{equation}

\section{Experiments}

\begin{table*}[!t]
\centering
\caption{\textbf{Performance comparison across varying transmission budgets.} We report Novel View Synthesis (NVS) and Novel Expression Synthesis (NES) using PSNR/SSIM/LPIPS. We also list the number of Gaussians, the amount of data to transmit (in Megabytes), and rendering speed. Rendering speed is measured on an RTX~4090 at $550\times 802$ resolution.}
\label{tab:lod_levels}
\vspace{4pt}
{\setlength{\tabcolsep}{4pt}%
\begin{tabular*}{\textwidth}{@{\extracolsep{\fill}} l ccc ccc r r c @{} }
\toprule
\multirow{2}{*}{} & \multicolumn{3}{c}{NVS} & \multicolumn{3}{c}{NES} & \multirow{2}{*}{\centering\arraybackslash \#Gaussians} & \multirow{2}{*}{\centering\arraybackslash Transmit Data} & \multirow{2}{*}{FPS} \\
\cmidrule(r){2-4} \cmidrule(lr){5-7}
 & PSNR$\uparrow$ & SSIM$\uparrow$ & LPIPS$\downarrow$ & PSNR$\uparrow$ & SSIM$\uparrow$ & LPIPS$\downarrow$ &  &  &  \\
\midrule
\mbox{GaussianAvatars} & 31.10 & 0.937 & 0.064 & 25.80 & 0.911 & 0.076 & 163,829 & 41.90 MB & 271 \\
5\% (Base) & 27.89 & 0.851 & 0.186 & 25.13 & 0.804 & 0.176 & 10,144  & 2.60 MB  & 291 \\
25\%       & 29.14 & 0.892 & 0.080 & 25.58 & 0.846 & 0.124 & 37,302  & 9.56 MB  & 278 \\
50\%       & 30.03 & 0.904 & 0.073 & 25.71 & 0.884 & 0.105 & 84,132  & 21.56 MB & 258 \\
100\%      & 31.47 & 0.929 & 0.068 & 25.89 & 0.908 & 0.080 & 169,438 & 43.42 MB & 260 \\
\bottomrule
\end{tabular*}}
\end{table*}

\begin{figure*}[htb]
    \centering
        \includegraphics[width=\textwidth]{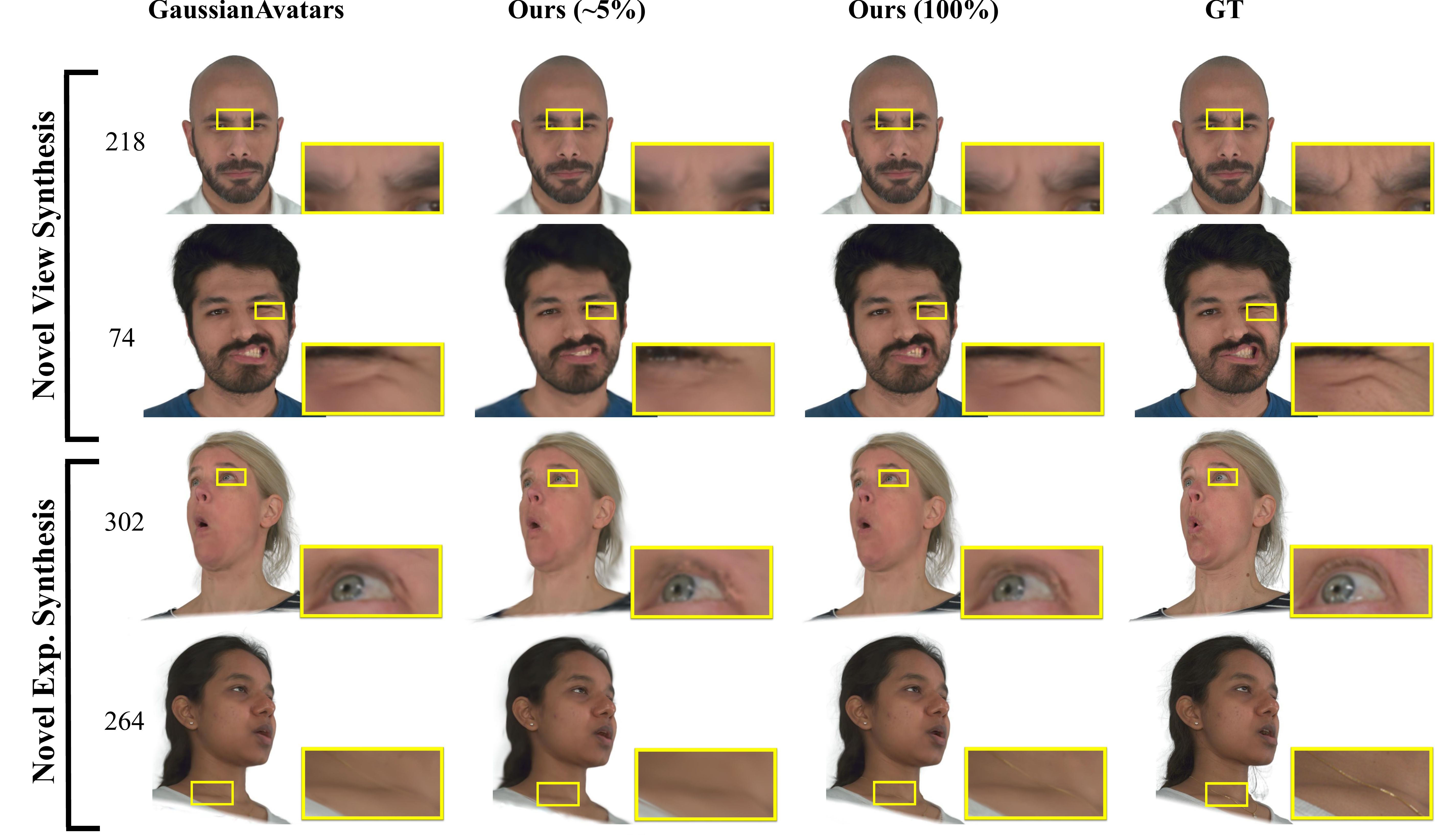}%
    \caption{\textbf{Qualitative comparison with state-of-the-art methods.}}
    \label{fig:com_result}
\end{figure*}

\begin{figure}[htb]
    \centering
        \includegraphics[width=\linewidth]{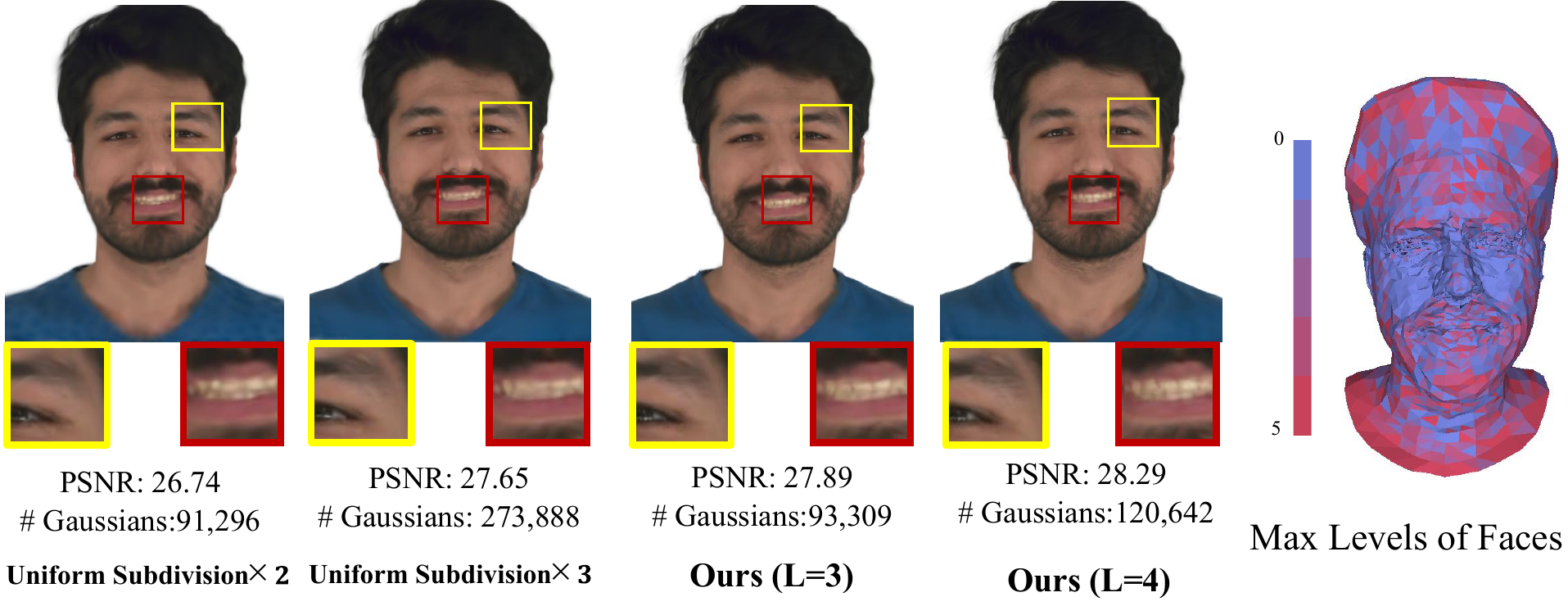}%
    \caption{\textbf{Comparison of adaptive and uniform subdivision.} Right: visualization of per-face subdivision levels. High-frequency regions like facial hair receive more aggressive splitting, whereas smoother areas require substantially fewer subdivisions.}
    \label{fig:adaptive_vs_uniform}
    
\end{figure}

We evaluate our method on the NeRSemble dataset~\cite{kirschstein2023nersemble}, which comprises multi-view videos for each subject with calibrated parameters for all 16 cameras. Following prior work~\cite{qian2024gaussianavatars,wang2025mega}, we downsample images to $802\times 550$ pixels and generate a foreground mask for each image. We adopt the same train/test split as GaussianAvatars~\cite{qian2024gaussianavatars}: 9 of the 10 expression sequences and 15 of the 16 cameras are used for training, with the remaining camera and the remaining expression sequence reserved for evaluation. All metrics are computed on pixels within the foreground mask.

\noindent \textbf{Implementation Details.}    We use the Adam optimizer~\cite{kingma2014adam} for all parameter updates. Learning rates are set to $1\times10^{-2}$ for the barycentric coordinates of subdivided points, $5\times10^{-3}$ for positions, $2\times10^{-2}$ for scales, and $1\times10^{-3}$ for rotations. The FLAME parameters vertex offsets, joint rotations, and expression coefficients are also optimized. We set $\lambda_{\text{pos}}=0.01$ and $\lambda_{\text{scale}}=1.0$. The maximum hierarchy level is $D=4$. Training runs for $60\mathrm{k}$ iterations, with the hierarchy expanded adaptively every $2\mathrm{k}$ iterations.

\subsection{Reconstruction and Animation Results}

We conduct comparative experiments with GaussianAvatars~\cite{qian2024gaussianavatars} and PointAvatar~\cite{Zheng_2023_CVPR}. All baselines are trained from scratch using their public implementations on the same training data, camera calibration, and preprocessing as ours. We compare reconstruction quality at our 100\% budget and 5\% base budget to state-of-the-art approaches. We report self-reenactment and novel view synthesis in terms of PSNR, SSIM, and LPIPS.

As shown in~\cref{fig:com_result}, our method reconstructs sharper details in several regions, particularly around the neck, shoulders, and clothing. These areas are relatively coarsely tessellated in the FLAME template compared with high‑saliency facial zones (e.g., the periocular region). Consequently, prior methods often allocate too few 3D Gaussians to these regions to faithfully capture their fine‑scale detail. In contrast, our adaptive growing strategy increases the number of Gaussians and refines the hierarchy only where needed, making allocation insensitive to FLAME’s non‑uniform tessellation.  Quantitative results in~\cref{tab:sota} show that our approach is on par with state‑of-the-art methods, and even at a minimal 5\% transmission budget, it yields a usable avatar suitable for bandwidth‑constrained streaming.

\subsection{Progressive Rendering Results}
We emulate a practical streaming setting in which the avatar is transmitted and rendered progressively. At inference, the renderer begins from the base structure (a 5\% data budget) and, within each structural level, activates refinement groups in descending order of a per‑face importance score until the next level is entered. This procedure induces a continuum of transmission percentages. For example, rendering at a 25\% budget uses all level‑1 Gaussians together with the top subset of level‑2 Gaussians by importance, while preserving all previously transmitted content. After every activation step, we evaluate novel‑view synthesis (NVS) and novel‑expression synthesis (NES), and we record the number of active Gaussians, the amount of data that must be transmitted, and the achieved frame rate (FPS). Rendering speed is measured on an RTX~4090 at $550\times 802$ resolution.~\cref{tab:lod_levels} summarizes these measurements together with image quality. For direct reference to a widely used baseline, we additionally report GaussianAvatars in the same table.

\begin{table}[ tb]
\centering
\footnotesize
\caption{\textbf{Quantitative comparison with SOTA methods.} We denote the \colorbox[HTML]{FD6864}{best} and \colorbox[HTML]{FFFC9E}{second best} scores in different colors.}
\label{tab:sota}
\vspace{4pt}
{\setlength{\tabcolsep}{3.5pt}%
\begin{tabularx}{\columnwidth}{@{} l *{3}{Y} *{3}{Y} @{} }
\toprule
& \multicolumn{3}{c}{NVS} & \multicolumn{3}{c}{NES} \\
\cmidrule(r){2-4} \cmidrule(l){5-7}
Method & PSNR$\uparrow$ & SSIM$\uparrow$ & LPIPS$\downarrow$ & PSNR$\uparrow$ & SSIM$\uparrow$ & LPIPS$\downarrow$ \\
\midrule
PointAvatar & 25.8 & 0.893 & 0.097 & 23.4 & 0.884 & 0.102 \\
GaussianAvatars & \second{31.1} & \best{0.937} & \best{0.064} & \second{25.8} & \best{0.911} & \best{0.076} \\
Ours (5\%) & 27.9 & 0.851 & 0.186 & 25.1 & 0.804 & 0.176 \\
Ours (100\%) & \best{31.5} & \second{0.929} & \second{0.068} & \best{25.9} & \second{0.908} & \second{0.080} \\
\bottomrule
\end{tabularx}}
\end{table}

~\cref{tab:lod_levels} exhibits monotonic improvements as additional data arrive. With only 2.60~MB transmitted (5\% budget), the avatar already attains reasonable quality. As higher-level Gaussians are streamed, fine structures such as shirt buttons, teeth, and hair gradually sharpen while temporal stability are maintained. At 100\% transmission, our approach achieves rendering quality comparable to SOTA methods. Notably, the frame rates do not drop significantly, likely because the 3DGS workload has not yet saturated the GPU. However,  multi-user VR scenarios can easily accumulate enough 3D Gaussians to hit GPU rasterization bottlenecks. In such computationally demanding cases, our method provides a critical advantage by flexibly balancing primitive count and visual fidelity

\begin{figure}[t]
    \centering
    \begin{minipage}[t]{0.53\columnwidth}
        \centering
        \includegraphics[width=\linewidth]{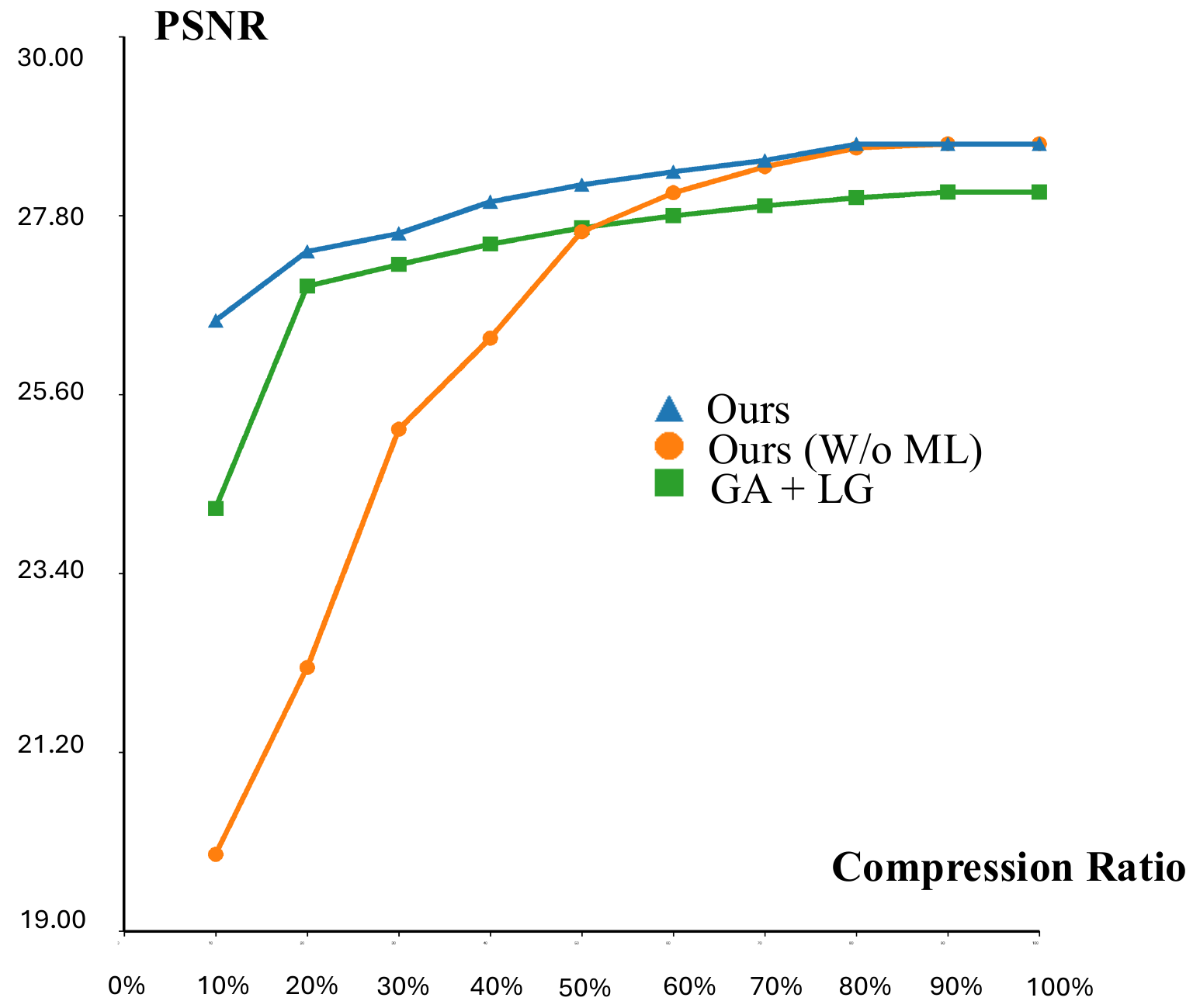}%
        \vspace{0.1em}
        \centerline{\small (a) PSNR-compression comparison}
    \end{minipage}\hfill
    \begin{minipage}[t]{0.45\columnwidth}
        \centering
        \includegraphics[width=\linewidth]{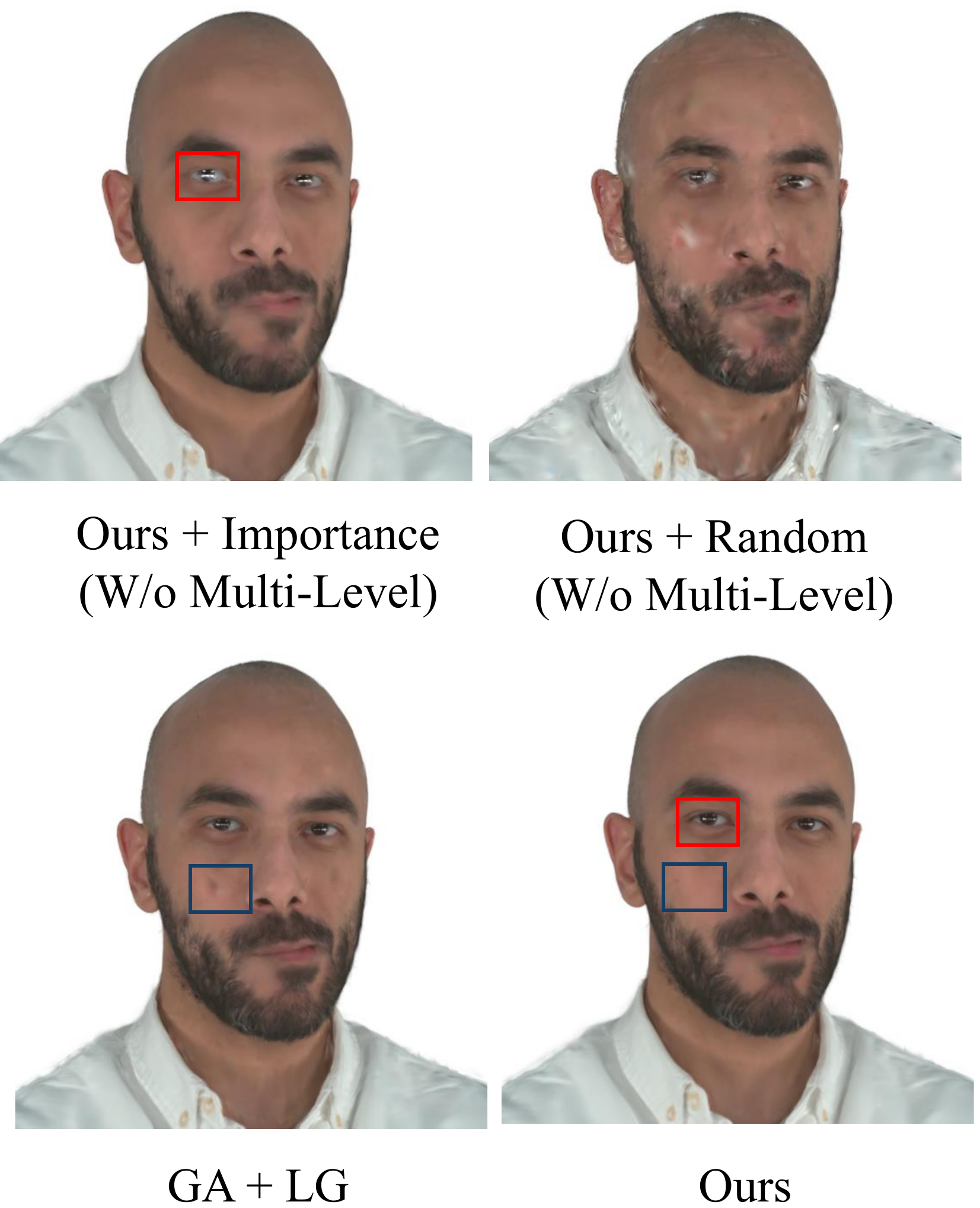}%
        \vspace{0.1em}
        \centerline{\small (b) Multi-level ablation}
    \end{minipage}
    \vspace{-0.5em}
\caption{{\textbf{Progressive streaming and multi-level ablation.}}}
    \label{fig:combined_analysis}
    \vspace{-0.8em}
\end{figure}

We further evaluate the continuous streaming capability by comparing our method against a discrete compression pipeline: GaussianAvatars combined with LightGaussian (GA+LG). As~\cref{fig:combined_analysis}(a) shows, GA+LG yields discrete operating points and demands 227.2~MB to store 10 levels. In contrast, our single progressive asset requires only 43.4~MB while supporting continuous, arbitrary-rate rendering and smooth quality refinement as data arrives.

\subsection{Ablation Study}

\noindent\textbf{Adaptive Growing.} We uniformly subdivide the initial mesh two and three times to highlight the importance of adaptive growing.  For the uniform baseline, we recursively subdivide the template mesh two or three times at the beginning of training and then disable adaptive growing. As shown in~\cref{fig:adaptive_vs_uniform}, compared with uniform subdivision, our adaptive growing strategy achieves higher reconstruction quality while using fewer 3D Gaussians.

\noindent\textbf{Multi-Level.} To verify the benefits of multi-level over importance ranking alone, we evaluate an importance-only variant at a 50\% transmission budget. As~\cref{fig:combined_analysis}(b) shows, relying solely on importance ranking causes local coverage holes. In contrast, our multi-level design guarantees global base-level coverage for smooth progressive rendering.

\begin{figure}[htb]
    \centering
        \includegraphics[width=\columnwidth]{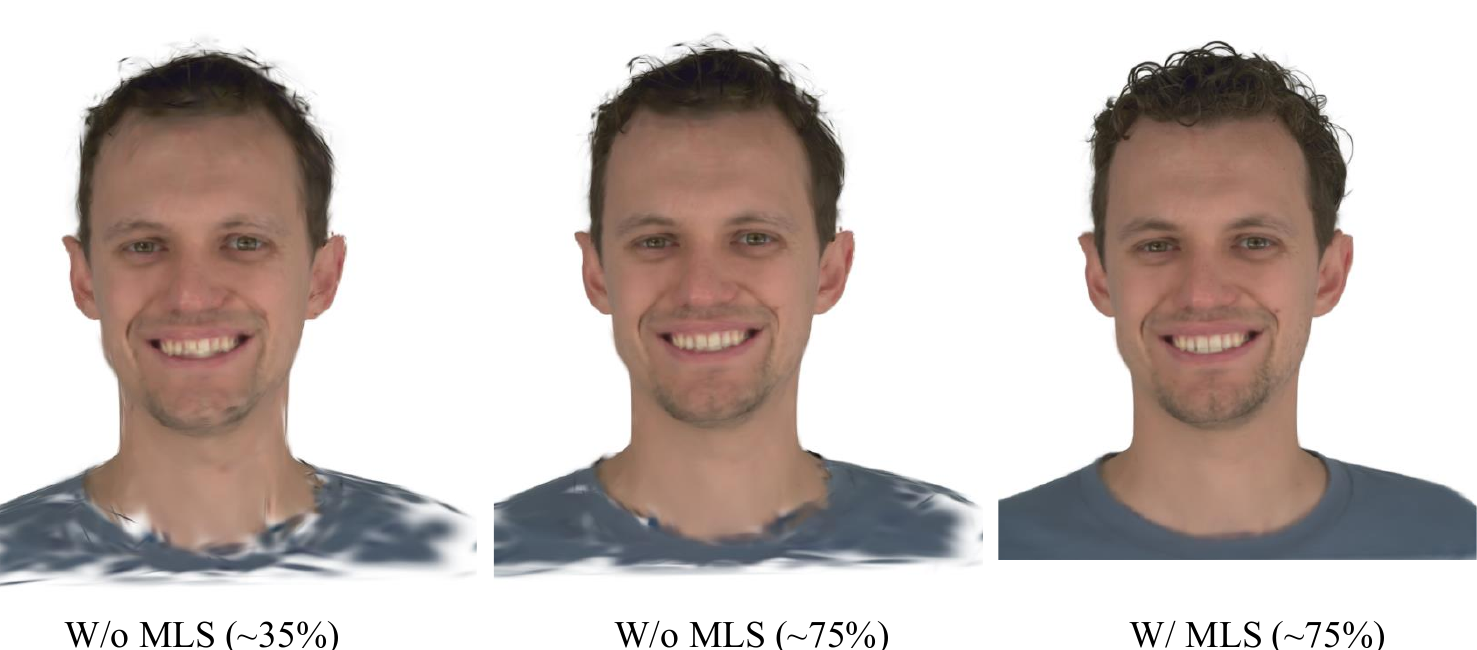}%
    \caption{\textbf{Ablation study on multi-level supervision.}}
    \label{fig:mls}
\end{figure}

\noindent\textbf{Supervision on Multi-Level.} As shown in~\cref{fig:mls} and ~\cref{tab:ablation_306}, supervising only the finest level while ignoring intermediate ones prevents lower levels from learning a complete head, which conflicts with our goal of quickly streaming a usable head under bandwidth constraints. In ``W/ Freeze'', we freeze the parameters of previous level every 50k iterations, optimizing only the newly added level. This stage-wise schedule forces high-resolution 3D Gaussians to reconstruct residual details on fixed, low-detail ones, reducing the model’s degrees of freedom and yielding lower quality than joint training where all levels are optimized.

\noindent\textbf{Importance Ranking.} In ``W/ Random'' of~\cref{tab:ablation_306}, we randomly permute the 3DGS within each level and load them in that order under the current bandwidth budget. Compared with randomly ordering the 3DGS at each level for progressive transmission and rendering, our importance ranking sorts Gaussians by their contribution to the rendered image, ensuring that high contribution Gaussians are loaded first and yielding finer rendering quality.

\begin{table}[tb]

\centering

\footnotesize

\caption{\textbf{Ablation study across all subjects.} We report average Novel View Synthesis (NVS) and Novel Expression Synthesis (NES) metrics over all subjects. ``W/o MLS'' supervises only the finest level, whereas ``W/ MLS'' supervises all levels.}

\label{tab:ablation_306}

{\setlength{\tabcolsep}{2pt}

\begin{tabular}{@{}p{0.18\columnwidth} c ccc ccc@{}}

\toprule

 & & \multicolumn{3}{c}{NVS} & \multicolumn{3}{c}{NES} \\

\cmidrule(r){3-5} \cmidrule(l){6-8}

Strategy & Budget & PSNR$\uparrow$ & SSIM$\uparrow$ & LPIPS$\downarrow$ & PSNR$\uparrow$ & SSIM$\uparrow$ & LPIPS$\downarrow$ \\

\midrule

W/o MLS              & 35\%   & 20.06 & 0.812 & 0.284 & 19.86 & 0.798 & 0.327 \\

W/ MLS        & 35\%   & 29.87 & 0.897 & 0.076 & 25.65 & 0.859 & 0.114  \\

\midrule

W/ Freeze            & 100\%  & 31.08 & 0.925 & 0.090 & 25.64 & 0.901 & 0.080 \\

W/ MLS      & 100\%  & 31.47 & 0.929 & 0.068 & 25.89 & 0.908 & 0.080 \\

\midrule

W/ Random            & 25\%   & 28.40 & 0.867 & 0.109 & 25.46 & 0.838 & 0.143 \\

W/ Ranking   & 25\%   & 29.14 & 0.892 & 0.080 & 25.58 & 0.846 & 0.124 \\

\bottomrule

\end{tabular}}

\end{table}
\section{Conclusion}

We proposed ProgressiveAvatars, a progressive, animatable 3D Gaussian avatar representation that enables progressive rendering of 3D avatars. By growing a mesh-anchored hierarchy via adaptive implicit subdivision and binding Gaussians in face-local frames, our representation supports incremental loading, progressive transmission, adaptive refinement, and progressive rendering. A usable avatar appears quickly and improves smoothly as additional data arrives, without replacing or deleting previously transmitted content. Experiments demonstrated strong time-to-quality under tight budgets and parity with state-of-the-art reconstruction at full detail. We believe ProgressiveAvatars plays an important role in progressive avatar streaming, a capability that will be vital for bringing 3DGS content to end users across heterogeneous networks and devices. While our work focuses on digital head avatars, the proposed progressive, mesh‑anchored Gaussian hierarchy is generic and can be readily extended to general‑purpose scenarios, such as progressive streaming and rendering of 3D assets and 4D volumetric video.

\section*{Acknowledgements}
\noindent This research was supported by the National Natural Science Foundation of China (No. 62272433), and the Fundamental Research Funds for the Central Universities.

{
    \small
    \bibliographystyle{ieeenat_fullname}
    \bibliography{main}
}


\end{document}